\documentclass[conference]{IEEEtran}
\IEEEoverridecommandlockouts
\usepackage{cite}
\usepackage{amsmath,amssymb,amsfonts}
\usepackage{algorithmic}
\usepackage{graphicx}
\usepackage{textcomp}
\usepackage{xcolor}
\usepackage{amsmath}
\usepackage{amssymb}
\usepackage{mathtools}
\usepackage{amsthm}
\usepackage{multirow}
\usepackage{ulem}
\usepackage{microtype}
\usepackage{graphicx}
\usepackage{subfigure}
\usepackage{booktabs} 

\usepackage{hyperref}

\def\BibTeX{{\rm B\kern-.05em{\sc i\kern-.025em b}\kern-.08em
    T\kern-.1667em\lower.7ex\hbox{E}\kern-.125emX}}
\begin{document}

\title{UnetTSF: A Better Performance Linear Complexity Time Series Prediction Model}

\author{\IEEEauthorblockN{1\textsuperscript{st} Chu  Li}
\IEEEauthorblockA{\textit{University of Science and Technology of China} \\
Hefei, China \\
lichuzm@mail.ustc.edu.cn}
\and
\IEEEauthorblockN{2\textsuperscript{nd} Bingjia Xiao}
\IEEEauthorblockA{\textit{Institute of Plasma Physics, Hefei Institutes of Physical Science} \\
\textit{Chinese Academy of Sciences}\\
Hefei, China \\
bjxiao@ipp.ac.cn}
\and
\IEEEauthorblockN{3\textsuperscript{rd} Qingping Yuan}
\IEEEauthorblockA{\textit{Institute of Plasma Physics, Hefei Institutes of Physical Science} \\
\textit{Chinese Academy of Sciences}\\
Hefei, China \\
qpyuan@ipp.ac.cn}

}

\maketitle

\begin{abstract}
	Recently, Transformer-base models have made significant progress in the field of time series prediction which have achieved good results and become baseline models beyond Dlinear. The paper proposes an U-Net time series prediction model (UnetTSF) with linear complexity, which adopts the U-Net architecture.We are the first to use FPN technology to extract features from time series data, replacing the method of decomposing time series data into trend and seasonal terms, while designing a fusion structure suitable for time series data. After testing on 8 open-source datasets, compared to the best linear model DLiner, Out of 32 testing projects, 31 achieved the best results,The average decrease in mse is 10.1\%, while the average decrease in mae is 9.1\%. Compared with the complex transformer-base PatchTST, UnetTSF obtained 9 optimal results for mse and 15 optimal results for mae in 32 testing projects.Code is available at https://github.com/lichuustc/UnetTSF.
\end{abstract}

\section{Introduction}
As is well known,time series prediction has always been one of the hot areas of deep learning research. time series prediction has important applications in transportation, energy, weather, finance, and other fields\cite{survey1}\cite{survey2}. In actual production and life, there is a lot of demand for long time series prediction. To meet the needs of production and life, researchers have launched various machine learning algorithms, and many excellent deep learning algorithms have emerged. From the traditional machine learning algorithm ARMA\cite{arima}, GBRT to recursive neural networks, causal time networks, etc.

Transformer\cite{Transformer} is currently one of the most successful sequence modeling architectures in the field of machine learning, exhibiting unparalleled performance in various applications such as natural language processing (NLP), speech recognition, and computer vision. In recent years, transformer-base time series prediction models have also emerged in large numbers and achieved good results, such as PatchTST\cite{PatchTST}, ETSformer, Autoformer\cite{autoformer}, FEDformer\cite{fedformer}, etc, However, Transformer-base models generally have the drawbacks of having multiple model parameters, high computational complexity, and long inference time. Therefore, Researchers have proposed a linear complexity model (DLlinear\cite{dlinear}). The DLinear\cite{dlinear} decomposes data into seasonal and trend terms, predicts them separately and adds them up. DLlinear\cite{dlinear} is a model composed of two fully connected layers, with extremely low model parameters and computational complexity. However, its prediction performance significantly surpasses many complex models such as Autoformer\cite{autoformer}, FEDformer\cite{fedformer}, and lightTST\cite{LightTS}, making it an important baseline model in the field of time series prediction.

\begin{figure*}[ht]
	\begin{center}
		\centerline{\includegraphics[scale=0.5]{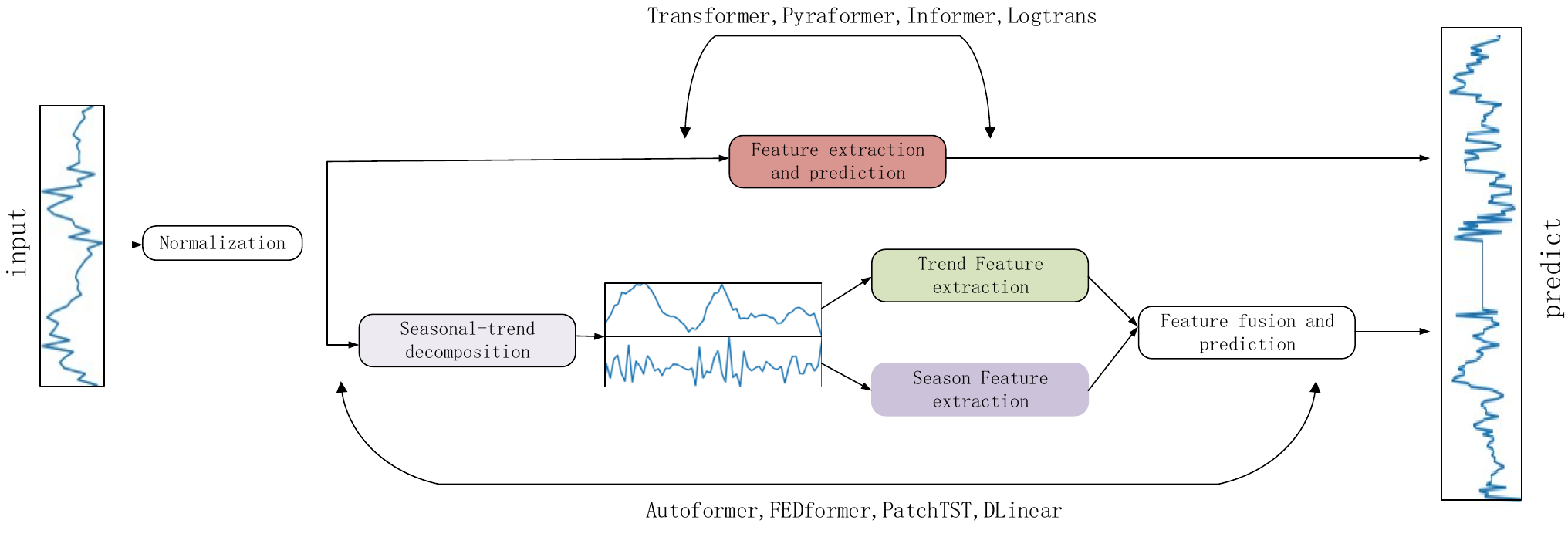}}
		\caption{The pipeline of existing Model TSF solutions.}
		\label{data-decompose}
	\end{center}
\end{figure*}

Preprocessing temporal data is one of the methods to improve the predictive ability of models, among which data decomposition and normalization are the most commonly used methods. The earliest origin of data decomposition is the ARMA\cite{arima}, which decomposes time series data into four parts: trend factor (T), cycle factor (C), seasonal factor (S), and random factor (I), which is more in line with human perception of data. Decomposing the data into trend and seasonal terms is a simplification of these four parts. As shown in Figure \ref{data-decompose},Traditional machine learning is based on raw data, extracting features from the raw data and predicting future data, such as GBRT, Informar\cite{informer}, Linear\cite{dlinear} and other models\cite{deepar}\cite{LSTMnetwork}. There is another data decomposition strategy: to decompose time series data into trend and seasonal items. The PatchTST\cite{PatchTST} and DLlinear\cite{dlinear} decompose time series data into trend and seasonal terms, extract and predict the two separately, and add them up to output the prediction results. Autoformer\cite{autoformer} and FEDformer\cite{fedformer} use feature fusion models to fuse trend and seasonal features, and use the fused features to predict the future. After verification by many scientific researchers, binary decomposition is a very suitable method for deep learning. However, the data binary decomposition method has two problems: 
\begin{itemize}
	\item The seasonal and trend characteristics of data items are fundamentally related. Simply and roughly decomposing the data into trend and seasonal items will cause the seasonal items to lose their trend characteristics, while the trend characteristics will also lose their seasonal characteristics
	\item  The trend and season terms enter their respective feature extraction modules for feature extraction and prediction, followed by prediction feature fusion. Currently, feature fusion mainly uses simple addition to output prediction results, and there is no correlation feature between the two for processing 
\end{itemize}

After DLiner\cite{dlinear}, transformer-base time series prediction models have once again become mainstream. However, these models suffer from high training resource consumption and slow inference speed. In order to solve these problems, this paper proposes a U-Net time series prediction model (UnetTSF). The main three contributions of this paper are as follows:
\begin{itemize}
	\item \textbf{Time series FPN method}: This article proposes an FPN\cite{FPN} method for describing time series data, using pooling functions to perform multi-layer operations on the data to extract trend information at different depths and jointly form a data group to replace the original data. Compared to the commonly used binary decomposition method in the field of temporal data, the FPN\cite{FPN} method uses small pooling kernels for data processing, which has lower computational complexity and is more effective in extracting temporal shallow and deep features. Shallow features include seasonal and trend terms, and as depth increases, seasonal features are gradually removed, retaining more trend features.
	\item \textbf{UnetTSF}:For the first time, we introduce the U-Net structure into the field of time series prediction. We combine the multi-level prediction characteristics of the U-Net network with time series data sets to predict data in the same level dimension. Through multi-step fusion, we integrate higher depth trend features into lower level features. The gradual fusion approach can better utilize the depth features of trend items, avoiding the influence of a large proportion of depth features on prediction results.
	\item The experimental results show that in terms of model parameter count and computational complexity. In the 32 tests of multivariate time series prediction, compared to DLiner\cite{dlinear}, UnetTSF achieved 31 optima in both mae and mse, with an average decrease of 10.1\% in mse and 9.1\% in mae. Compared with PatchTST\cite{PatchTST}, UnetTSF achieved 9 optima in terms of mse and 15 optima in terms of mae.
\end{itemize}
Below, we will provide a detailed introduction to our method and demonstrate its effectiveness through extensive experiments.
\begin{figure*}[ht]
	\begin{center}
		\centerline{\includegraphics[scale=0.8]{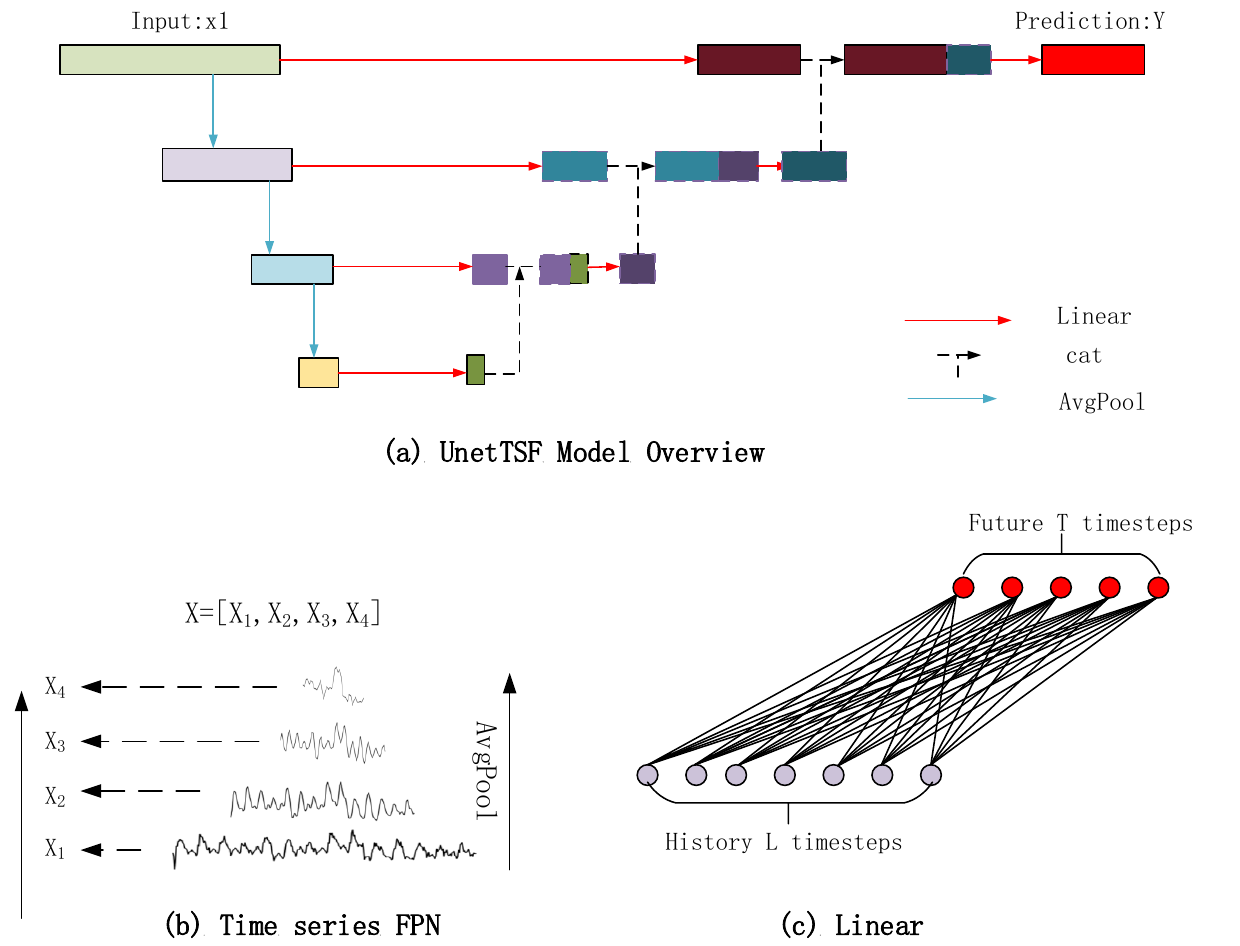}}
		\caption{UnetTSF architecture.(a)The overall inference process of UnetTSF model.The model consists of downsampling and fully connected components, with linear complexity.(b)Using avgpool to complete FPN sampling of temporal data.(c)Illustration of the basic linear model.}
		\label{UnetTSF architecture}
	\end{center}
\end{figure*}
\section{Related work}
\textbf{Time series linear models}: time series linear models refer to models with a complexity of $O (L)$, mainly including ARMA, NLinear, DLlinear, etc. The DLlinear\cite{dlinear} model and NLlinear are currently the fundamental models in the long-term time series forecasting task. Linear is the normalization of input data into a fully connected network layer. NLlinear model uses the nearest value of $x_L$ as the basis and $x-x_L$ to construct a new historical data $x^{'}$, enter a layer of fully connected output prediction result $y^{'}$, and add the predicted result $y^{'}$ to $x_L$ to output the predicted data $y$. The DLlinear\cite{dlinear} model divides time series data into seasonal-term $x_{season}$ and trend-term $x_{trend}$ through data decomposition, and predicts seasonal term $y_{season}$ and trend term $y_{trend}$ using fully connected layers respectively. The two are added together to obtain the predicted data $y$. The complexity of a linear model is $ O (L)$.

\textbf{Transformer models}:With the breakthrough progress of Transformer in many fields such as NLP\cite{bert} and imaging, a large number of researchers have also applied Transformer models to long-term time series prediction. For example, LogTrans\cite{logtrans} uses convolutional self attention layers and LogSparse design to capture local information and reduce spatial complexity. Informer\cite{informer} proposed a ProbSparse self attention extraction technique that can effectively extract the most important features. Autoformer\cite{autoformer} draws inspiration from traditional time series analysis methods to decompose time series data into trend and seasonal terms, and proposes the concept of autocorrelation. FEDformer\cite{fedformer} uses Fourier enhanced structures to achieve linear complexity. Pyraformer proposed a pyramid attention module with inter scale and intra scale connections, which also achieved linear complexity. PatchTST\cite{PatchTST} applies patch technology to shorten sequence length, significantly reduce model complexity, and enhance local features of the representation sequence.

\textbf{Time series decomposition}: A time typically consists of long-term trends, seasonal fluctuations, cyclical fluctuations, and irregular fluctuations. Long term trend refers to a trend or state in which a phenomenon continues to develop and change over a long period of time. Seasonal fluctuations are regular changes in the development level of phenomena caused by seasonal changes. Cyclic fluctuations refer to periodic and continuous changes that do not follow strict rules during a certain period of time. Irregular fluctuations refer to the impact of numerous accidental factors on a time series. Autoformer\cite{autoformer} first applies seasonal trend decomposition, which is time series analysis to make the raw data more predictable. Specifically, they input the trend period components extracted from the sequence. The difference between the original sequence and trend components is considered seasonal. Based on Autoformer's decomposition scheme, FEDformer\cite{fedformer} proposes an expert strategy for mixed extraction of trend components. By moving average kernels with various kernel sizes, PatchTST\cite{PatchTST} and DLinear\cite{dlinear} both use seasonal trend binary decomposition methods to separately predict seasons and trend items, and then add them up to obtain the prediction results

\textbf{U-Net}:The U-Net\cite{Unet} architecture, also known as ResNet in the field of image segmentation, is a well-known network architecture in the field of medical imaging. It performs well on small training datasets. The overall structure of U-Net\cite{Unet} is concise, stable, and efficient, and the strong compatibility of Encoder Decoder structure allows U-Net\cite{Unet} to seamlessly integrate with new generation models such as Transformer in both segmentation and generation fields. The compression characteristics of the Encoder module in U-Net\cite{Unet}. As the initial application of the Encoder module, the input image is downsampled to extract high-dimensional features that are much smaller than the original image, which is equivalent to performing compression operations. The Decoder module in U-Net utilizes multi-layer gradual fusion operations to effectively denoise and preserve effective features.

\section{Proposed Method}
The time series prediction problem is to predict future data with a length of {T} given historical data with an input length of {L}.Input historical data $ X=\{x_1, x_2, x_3,... ,x_L\} $ with a look-back window of fixed length L , the model outputs predicted data $x =\{x_{L+1},x_{L+2},...,x_{L+T}\} $, where $ x_i $ represents a vector dimension of $C$ at time $t=i$, and $C$ represents the number of channels in the input dataset.We designed UnetTSF using the U-Net\cite{Unet}  architecture and specifically designed time series FPN\cite{FPN} and multi-step fusion modules suitable for temporal data, as shown in Figure \ref{UnetTSF architecture}

\textbf{UnetTSF model}:UnetTSF consists of fully connected layers and pooling layers. The left side of the model mainly consists of time series FPN, and the pooling function is used to form the descriptive feature $ X=\{X_1, X_2, X_3,... ,X_{stage}\} $ of the input data. The stage represents the number of layers of the Unet network, and the right side of the model is the fusion module. The fully connected layer is used to fuse the upper layer features with the local layer features to output the final feature of the current layer, while the feature length remains unchanged.

\textbf{Time series FPN}:data decomposition is often used by time series prediction models to extract features from time series data. Generally, the data is divided into seasonal, periodic, trend, and fluctuation terms. Autoformer and DLiner both use large-scale adaptive smoothing kernels to extract trend terms from the original data. Subtracting the trend term from the original data results in seasonal terms, which can cause certain feature loss. Therefore, we adopt a multi-level extraction approach, For example, setting the data to be divided into 4 layers($ stage=4 $) , using the avgpool  Extract trend features with a configuration of $ kernel\_szie=3 ,stride=2 $ and $ padding=0 $, set the original input data as $ x $, and after passing through the FPN module, form four levels of input data:$ X  = [ x_1, x_2, x_3, x_4] $.
\[ x_1 = x \]
\[ x_2 = AvgPool(x_1) \]
\[ x_3 = AvgPool(x_2) \]
\[ x_4 = AvgPool(x_3) \]
\[{\rm{len(}}{{\rm{x}}_i}{\rm{)  = }}\left\lfloor {\frac{{{x_{i - 1}} + 2 \times padding - \ker nel\_size}}{{stride}} + 1} \right\rfloor \]
As shown in (b) of Figure \ref{UnetTSF architecture}, the FPN structure of temporal data can effectively extract trend features. The trend information in the top layer of the pyramid is more concentrated than in the bottom layer, and the seasonal features in the bottom layer of the pyramid are more abundant.

\textbf{Multi stage fusion module}: Through the temporal data FPN module, a multi-scale temporal feature $ X $ is formed. To fully utilize these features, multiple fully connected predictions are used to obtain $ Y = [y_1, y_2, y_3, y_4] $. The length calculation method of $ y_i $ is the same as $ x $, and the same pooling operation is used to calculate the length at each level as X. The fusion module adopts $ y_i $ and $ y_{i-1} $  is spliced,, and then a fully connected layer is used to output $ {y_i}^{'} $, The length of $  {y_i}^{'} $ and $ y_i $  is the same.
\[  len(y1) = len(y) = T \]
\[y_{i - 1}^{'} = Linear(cat({y_i^{'} },{y_{i - 1}}))\]
\[{\rm{len(}}{{\rm{y}}_i}{\rm{)  = }}\left\lfloor {\frac{{{y_{i - 1}} + 2 \times padding - \ker nel\_size}}{{stride}} + 1} \right\rfloor \]

\begin{table*}[htbp]
	\caption{The statistics of the nine popular datasets for benchmark.}
	\label{tab:dataset}
	\vskip 0.15in
	\begin{center}
		\begin{small}
			\begin{tabular}{l|ccccccccr}
				\toprule
				Datasets & ETTh1 & ETTh2 & ETTm1 & ETTm2 & traffic & Electricity & Weather & ILI  \\
				\midrule
				Variates   & 7 & 7 & 7 & 7& 862 &321 &211 &7  \\
				Timessteps & 17420 & 17420 & 69680 & 69680 & 17544 & 26304  &52696 &966\\
				Granularity & 1hour & 1hour & 5min & 5min & 1hour & 1hour  &10min &11week\\
				\bottomrule
			\end{tabular}
		\end{small}
	\end{center}
	\vskip -0.1in
\end{table*}

\section{Experiments}

\begin{table*}[htbp]
	\caption{Multivariate long-term forecasting results. We use prediction lengths $T\in \{24, 36, 48, 60\}$ for ILI dataset and $T\in \{96, 192, 336, 720\}$ for the others. The best results are in \textbf{bold} and the second best are \uline{underlined}.}
	\label{tab:multivarite}
	\vskip 0.15in
	\centering
	\resizebox{\linewidth}{!}{
		\begin{tabular}{cc|c|cc|cc|cc|cc|cc|cc|cc}
			\cline{2-17}
			&\multicolumn{2}{c|}{Models}& \multicolumn{2}{c|}{UnetTSF}& \multicolumn{2}{c|}{PatchTST}&  \multicolumn{2}{c|}{DLinear}& \multicolumn{2}{c|}{FEDformer}& \multicolumn{2}{c|}{Autoformer}& \multicolumn{2}{c|}{Informer}& \multicolumn{2}{c}{Pyraformer} \\
			\cline{2-17}
			&\multicolumn{2}{c|}{Metric}&MSE&MAE&MSE&MAE&MSE&MAE&MSE&MAE&MSE&MAE&MSE&MAE&MSE&MAE\\
			\cline{2-17}
			&\multirow{4}*{\rotatebox{90}{Weather}}& 96    & \textbf{0.145} & \uline{0.210}& \uline{0.147} & \textbf{0.198} & 0.176 & 0.237 & 0.238 & 0.314 & 0.249 & 0.329 & 0.354 & 0.405 & 0.896 & 0.556  \\
			&\multicolumn{1}{c|}{}& 192   & \textbf{0.187} & \textbf{0.234} &  \uline{0.190} & \uline{0.240} & 0.220 & 0.282 & 0.275 & 0.329 & 0.325 & 0.370 & 0.419 & 0.434 & 0.622 & 0.624 \\
			&\multicolumn{1}{c|}{}& 336   & \textbf{0.238} & \textbf{0.274} & \uline{0.242} & \uline{0.282} & 0.265 & 0.319 & 0.339 & 0.377 & 0.351 & 0.391 & 0.583 & 0.543  & 0.739 & 0.753 \\
			&\multicolumn{1}{c|}{}& 720   & \textbf{0.304}& \textbf{0.325} & \textbf{0.304} & \uline{0.328} & 0.323 & 0.362 & 0.389 & 0.409 & 0.415 & 0.426 & 0.916 & 0.705 & 1.004 & 0.934 \\
			\cline{2-17}
			&\multirow{4}*{\rotatebox{90}{Traffic}}& 96    & \uline{0.395} & \uline{0.263} &  \textbf{0.360} & \textbf{0.249} & 0.410 & 0.282 & 0.576 & 0.359 & 0.597 & 0.371 & 0.733 & 0.410 & 2.085 & 0.468  \\
			&\multicolumn{1}{c|}{} & 192  & \uline{0.406} & \uline{0.277}  & \textbf{0.379} & \textbf{0.256} & 0.423 & 0.287 & 0.610 & 0.380 & 0.607 & 0.382 & 0.777 & 0.435 & 0.867 & 0.467 \\
			&\multicolumn{1}{c|}{}& 336  & \uline{0.422} & \uline{0.285} &  \textbf{0.392} & \textbf{0.264} & 0.436 & 0.296 & 0.608 & 0.375 & 0.623 & 0.387 & 0.776 & 0.434 & 0.869 & 0.469  \\
			&\multicolumn{1}{c|}{}& 720  & \uline{0.443} & \uline{0.299} &  \textbf{0.432} & \textbf{0.286} & 0.466 & 0.315 & 0.621 & 0.375 & 0.639 & 0.395 & 0.827 & 0.466 & 0.881 & 0.473  \\
			\cline{2-17}
			&\multirow{4}*{\rotatebox{90}{Electricity}}& 96   & \uline{0.132} & \uline{0.229} &  \textbf{0.129} & \textbf{0.222} & 0.140 & 0.237 & 0.186 & 0.302 & 0.196 & 0.313 & 0.304 & 0.393 & 0.386 & 0.449  \\
			&\multicolumn{1}{c|}{}& 192  & \uline{0.146} & \uline{0.244} & \textbf{0.141} & \textbf{0.241} & 0.153 & 0.249 & 0.197 & 0.311 & 0.211 & 0.324 & 0.327 & 0.417 & 0.386 & 0.443  \\
			&\multicolumn{1}{c|}{}& 336  & \textbf{0.162} & \uline{0.262}  & \uline{0.163} & \textbf{0.259} & 0.169 & 0.267 & 0.213 & 0.328 & 0.214 & 0.327 & 0.333 & 0.422 & 0.378 & 0.443  \\
			&\multicolumn{1}{c|}{}& 720  & \uline{0.200} & \uline{0.297} &  \textbf{0.197} & \textbf{0.290} & 0.203 & 0.301 & 0.233 & 0.344 & 0.236 & 0.342 & 0.351 & 0.427 & 0.376 & 0.445  \\
			\cline{2-17}
			&\multirow{4}*{\rotatebox{90}{ILI}}& 24   & \uline{1.696} & \uline{0.789} &  \textbf{1.281} & \textbf{0.704} & 2.215 & 1.081 & 2.624 & 1.095 & 2.906 & 1.182 & 4.657 & 1.449  & 1.420 & 2.012  \\
			&\multicolumn{1}{c|}{} & 36    & \uline{1.693} & \uline{0.811}  & \textbf{1.251}& \textbf{0.752} & 1.963 & 0.963 & 2.516 & 1.021 & 2.585 & 1.038 & 4.650 & 1.463 & 7.394 & 2.031\\
			&\multicolumn{1}{c|}{}& 48   & \uline{1.867} & \uline{0.881} & \textbf{1.673} &\textbf{0.854} & 2.130 & 1.024 & 2.505 & 1.041 & 3.024 & 1.145 & 5.004 & 1.542 & 7.551 & 2.057  \\
			&\multicolumn{1}{c|}{}& 60   & \textbf{1.421} & \textbf{0.747}  & \uline{1.526} & \uline{0.795} & 2.368 & 1.096 & 2.742 & 1.122 & 2.761 & 1.114 & 5.071 & 1.543 & 7.662 & 2.100 \\
			\cline{2-17}
			&\multirow{4}*{\rotatebox{90}{ETTh1}}& 96   & \textbf{0.368} & \textbf{0.394} & \uline{0.370} & {0.400} & {0.375} & \uline{0.399} & 0.376 & 0.415 & 0.435 & 0.446 & 0.941 & 0.769 & 0.664 & 0.612  \\
			&\multicolumn{1}{c|}{}& 192  & \uline{0.406} & \uline{0.417} & {0.413} & {0.431} & \textbf{0.405} & \textbf{0.416} & 0.423 & 0.446 & 0.456 & 0.457 & 1.007 & 0.786 & 0.790 & 0.681  \\
			&\multicolumn{1}{c|}{}& 336  & \textbf{0.408} & \textbf{0.425} & \uline{0.422} & \uline{0.440} & 0.439 & 0.443 & 0.444 & 0.462 & 0.486 & 0.487 & 1.038 & 0.784 & 0.891 & 0.738  \\
			&\multicolumn{1}{c|}{}& 720  & \uline{0.458} & \textbf{0.462}  & \textbf{0.447} &  \textbf{0.468} & 0.472 & 0.490 & 0.469 & 0.492 & 0.515 & 0.517 & 1.144 & 0.857 & 0.963 & 0.782 \\
			\cline{2-17}
			&\multirow{4}*{\rotatebox{90}{ETTh2}}& 96  & \uline{0.279} & \textbf{0.333}   & \textbf{0.274} & \uline{0.336} & 0.289 & 0.353 & 0.332 & 0.374 & 0.332 & 0.368 & 1.549 & 0.952 & 0.645 & 0.597\\
			&\multicolumn{1}{c|}{}& 192  & \uline{0.343} & \uline{0.395} &  \textbf{0.339} & \textbf{0.379} & 0.383 & 0.418 & 0.407 & 0.446 & 0.426 & 0.434 & 3.792 & 1.542 & 0.788 & 0.683\\
			&\multicolumn{1}{c|}{}& 336  & \uline{0.379} & \uline{0.423}  & \uline{0.331} & \textbf{0.380} & 0.448 & 0.465 & 0.400 & 0.447 & 0.477 & 0.479 & 4.215 & 1.642 & 0.907 & 0.747\\
			&\multicolumn{1}{c|}{}& 720   & \uline{0.446} & \uline{0.464}   & \textbf{0.379} & \textbf{0.422} & 0.605 & 0.551 & 0.412 & 0.469 & 0.453 & 0.490 & 3.656 & 1.619 & 0.963 & 0.783\\
			\cline{2-17}
			&\multirow{4}*{\rotatebox{90}{ETTm1}}& 96   & \textbf{0.287} & \textbf{0.336} &\uline{0.290} & \uline{0.342} & 0.299 & {0.343} & 0.326 & 0.390 & 0.510 & 0.492 & 0.626 & 0.560 & 0.543 & 0.510  \\
			&\multicolumn{1}{c|}{}& 192  & \uline{0.330} & \textbf{0.359}  & \textbf{0.328} & 0.365 & \uline{0.335} &{0.365} & 0.365 & 0.415 & 0.514 & 0.495 & 0.725 & 0.619 & 0.557 & 0.537 \\
			&\multicolumn{1}{c|}{}& 336  & \uline{0.368} & \textbf{0.380} & \textbf{0.361} & {0.393} & \uline{0.369} & {0.386} & 0.392 & 0.425 & {0.510} & {0.492} & 1.005 & 0.741 & 0.754 & 0.655  \\
			&\multicolumn{1}{c|}{}& 720  & \uline{0.425} & \textbf{0.413} & \uline{0.416} & 0.419 & 0.425 & {0.421} & 0.446 & 0.458 & 0.527 & 0.493 & 1.133 & 0.845 & 0.908 & 0.724  \\
			\cline{2-17}
			&\multirow{4}*{\rotatebox{90}{ETTm2}} & 96  & \uline{0.163} & \textbf{0.250}  & \textbf{0.162} & \uline{0.254} & 0.167 & 0.260 & 0.180 & 0.271 & 0.205 & 0.293 & 0.355 & 0.462& 0.435 & 0.507  \\
			&\multicolumn{1}{c|}{}& 192  & \textbf{0.216} & \textbf{0.287} & \uline{0.216} & \uline{0.293} & 0.224 & 0.303 & 0.252 & 0.318 & 0.278 & 0.336 & 0.595 & 0.586 & 0.730 & 0.673  \\
			&\multicolumn{1}{c|}{}& 336  & \uline{0.271} & \textbf{0.324}  & \textbf{0.269} & \uline{0.329} & 0.281 & 0.342 & 0.324 & 0.364 & 0.343 & 0.379 & 1.270 & 0.871 & 1.201 & 0.845  \\
			&\multicolumn{1}{c|}{}& 720 & \uline{0.360} & \uline{0.389}  & \textbf{0.350} & \textbf{0.380} & 0.397 & 0.421 & 0.410 & 0.420 & 0.414 & 0.419 & 3.001 & 1.267 & 3.625 & 1.451  \\
			\cline{2-17}
		\end{tabular}
	}
	\vskip -0.15in
\end{table*}

\begin{table*}[htbp]
	\caption{Univariate long-term forecasting results. ETT datasets are used with prediction lengths $T\in \{96, 192, 336, 720\}$. The best results are in \textbf{bold}.}
	\centering
	\label{tab::univariate}
	\vskip 0.15in
	\centering
	\resizebox{\linewidth}{!}{
		\begin{tabular}{cc|c|cc|cc|cc|cc|cc|cc|ccc}
			\cline{2-17}
			&\multicolumn{2}{c|}{Models}& \multicolumn{2}{c|}{UnetTSF}& \multicolumn{2}{c|}{PatchTST}& \multicolumn{2}{c|}{DLinear}& \multicolumn{2}{c|}{FEDformer}& \multicolumn{2}{c|}{Autoformer}& \multicolumn{2}{c|}{Informer}& \multicolumn{2}{c}{LogTrans}& \\
			\cline{2-17}
			&\multicolumn{2}{c|}{Metric}&MSE&MAE&MSE&MAE&MSE&MAE&MSE&MAE&MSE&MAE&MSE&MAE&MSE&MAE\\
			\cline{2-17}
			&\multirow{4}*{\rotatebox{90}{ETTh1}}& 96    & \textbf{0.055} & \textbf{0.178} & \textbf{0.055} & {0.179} & 0.056 & 0.180 & 0.079 & 0.215 & 0.071 & 0.206 & 0.193 & 0.377 & 0.283 & 0.468 \\
			&\multicolumn{1}{c|}{}& 192   & \textbf{0.071} & 0.205 & \textbf{0.071} & 0.205 & \textbf{0.071} & \textbf{0.204} & 0.104 & 0.245 & 0.114 & 0.262 & 0.217 & 0.395 & 0.234 & 0.409 \\
			&\multicolumn{1}{c|}{}& 336   & {0.078} & {0.221} & \textbf{0.076} & \textbf{0.220} & 0.098 & 0.244 & 0.119 & 0.270 & 0.107 & 0.258 & 0.202 & 0.381 & 0.386 & 0.546 \\
			&\multicolumn{1}{c|}{}& 720   & \textbf{0.087} & 0.233 & \textbf{0.087} & \textbf{0.232} & 0.189 & 0.359 & 0.142 & 0.299 & 0.126 & 0.283 & 0.183 & 0.355 & 0.475 & 0.629 \\
			\cline{2-17}
			&\multirow{4}*{\rotatebox{90}{ETTh2}}& 96    & \textbf{0.124} & \textbf{0.271} & {0.129} & 0.282 & 0.131 & {0.279} & 0.128 & 0.271 & 0.153 & 0.306 & 0.213 & 0.373 & 0.217 & 0.379 \\
			&\multicolumn{1}{c|}{}& 192   & \textbf{0.165} & \textbf{0.320} & {0.168} & {0.328} & 0.176 & 0.329 & 0.185 & 0.330 & 0.204 & 0.351 & 0.227 & 0.387 & 0.281 & 0.429 \\
			&\multicolumn{1}{c|}{}& 336   & {0.184} & {0.348} & \textbf{0.171} & \textbf{0.336} & 0.209 & 0.367 & 0.231 & 0.378 & 0.246 & 0.389 & 0.242 & 0.401 & 0.293 & 0.437 \\
			&\multicolumn{1}{c|}{}& 720   & {0.227} & {0.382} & \textbf{0.223} & \textbf{0.380} & 0.276 & 0.426 & 0.278 & 0.420 & 0.268 & 0.409 & 0.291 & 0.439 & 0.218 & 0.387 \\
			\cline{2-17}
			&\multirow{4}*{\rotatebox{90}{ETTm1}}& 96    & \textbf{0.026} & \textbf{0.121} & \textbf{0.026} & \textbf{0.121} & 0.028 & 0.123 & 0.033 & 0.140 & 0.056 & 0.183 & 0.109 & 0.277 & 0.049 & 0.171 \\
			&\multicolumn{1}{c|}{}& 192   & \textbf{0.039} & \textbf{0.149} & \textbf{0.039} & {0.150} & 0.045 & 0.156 & 0.058 & 0.186 & 0.081 & 0.216 & 0.151 & 0.310 & 0.157 & 0.317 \\
			&\multicolumn{1}{c|}{}& 336   & \textbf{0.051} & \textbf{0.171} & \textbf{0.053} & {0.173} & 0.061 & 0.182 & 0.084 & 0.231 & 0.076 & 0.218 & 0.427 & 0.591 & 0.289 & 0.459 \\
			&\multicolumn{1}{c|}{}& 720   & \textbf{0.071} & \textbf{0.203} & {0.073} & {0.206} & 0.080 & 0.210 & 0.102 & 0.250 & 0.110 & 0.267 & 0.438 & 0.586 & 0.430 & 0.579 \\
			\cline{2-17}
			&\multirow{4}*{\rotatebox{90}{ETTm2}}& 96    & \textbf{0.063} & \textbf{0.181} & 0.065 & 0.186 & \textbf{0.063} & {0.183} & 0.067 & 0.198 & 0.065 & 0.189 & 0.088 & 0.225 & 0.075 & 0.208 \\
			&\multicolumn{1}{c|}{}& 192   & \textbf{0.090} & \textbf{0.224} & 0.094 & 0.231 & {0.092} & {0.227} & 0.102 & 0.245 & 0.118 & 0.256 & 0.132 & 0.283 & 0.129 & 0.275 \\
			&\multicolumn{1}{c|}{}& 336   & \textbf{0.116} & \textbf{0.256} & 0.120 & 0.265 & {0.119} & {0.261} & 0.130 & 0.279 & 0.154 & 0.305 & 0.180 & 0.336 & 0.154 & 0.302 \\
			&\multicolumn{1}{c|}{}& 720   & \textbf{0.170} & \textbf{0.318} & {0.171} & 0.322 & 0.175 & {0.320} & 0.178 & 0.325 & 0.182 & 0.335 & 0.300 & 0.435 & 0.160 & 0.321 \\
			\cline{2-17}
			&\multicolumn{2}{c|}{Count}&13&11&9&4&2&1&0&0&0&0&0&0&0&0\\
			\cline{2-17}
		\end{tabular}
	}
	\vskip -0.15in
\end{table*}
\textbf{Dataset}: We evaluate the performance of our proposed UnetTSF on eight widely used real-world datasets, including ETT (ETTh1, ETTh2, ETTm1, ETTm2), transportation, electricity, weather, and ILI.These datasets have been extensively utilized for benchmarking and publicly available on \cite{autoformer}. We would like to emphasize several large datasets: weather, transportation, ILI, and electricity. They have more time series, and compared to other smaller datasets, the results will be more stable and less susceptible to overfitting. Univariate time series prediction testing will be conducted on the ETT dataset, while multivariate time series prediction testing will be conducted on 8 datasets. Table 1 summarizes the statistical data of these datasets.The statistics of those datasets are summarized In Table \ref{tab:dataset}.

\textbf{Evaluation metric}: Following previous works, we use Mean Squared Error (MSE) and Mean Absolute
Error (MAE)\cite{mae} as the core metrics to compare performance.

\textbf{Compared SOTA methods}: We choose the SOTA Transformer-based models:PatchTST\cite{PatchTST}, FEDformer\cite{fedformer}, Autoformer\cite{autoformer}, Informer\cite{informer}.At the same time, we choose the best linear model DLiner\cite{dlinear} as the baseline model.All of the models are following the same experimental setup with prediction length $ T \in [24,36,48,60]  $ for ILI dataset and $ T \in [96, 192, 336, 720]  $ for other datasets as in the original papers. We get baseline results of models from PachTST\cite{PatchTST} and DLinear\cite{dlinear} with the default look-back window $L \le 720 $.The result of patchTST\cite{PatchTST} is taken from the optimal results of patchTST/64 and patchTST/42.

\textbf{Platform}:UnetTSF was trained/tested on a single Nvidia RTX A4000 16GB GPU.
\section{Results and Analysis}
\begin{table}[htbp]
	\caption{Compare the parameter quantity and resource consumption of the model under the look-back window $ L = 336 $ and $  T = 96  $ on the ETTh2.MACS are the number of multiply-accumulate operations.}
	\label{tab:parameter}
	\vskip 0.15in
	\begin{center}
		\begin{small}
			\begin{tabular}{l|ccc}
				\toprule
				Method & MACs & Parameter  & Memory  \\
				\midrule
				UnetTSF   & 13.56M &  0.42M  & 20.39M  \\
				\hline
				DLinear & 14.52M & 0.45M  & 20.55M \\
				\hline
				PatchTST & 164.97M & 0.46M  & 20.72M \\
				\hline
				Autoformer & 90517.61M & 10.54M  & 119.39M \\
				\hline
				Informer & 79438.08M & 11.33M  & 126.22M \\
				\bottomrule
			\end{tabular}
		\end{small}
	\end{center}
	\vskip -0.1in
\end{table}

\textbf{Univarite Time-series Forecasting}:Table \ref{tab::univariate} summarize the univariate evaluation results of all the methods on ETT datasets, which is the univariate series that we are trying to forecast. We cite the baseline results from \cite{PatchTST} and \cite{dlinear}.Among the 16 tests on the ETT dataset, overall, UnetTSF achieved 13 optima in terms of mse and 11 optima in terms of mae. Compared with DLiner, UnetTSF achieved a maximum reduction of 54\% in mse, an average decrease of 11.0\%, a maximum reduction of 35.0\% in mae, and an average decrease of 5.5\%. Compared with PatchTST, UnetTSF has 7 leading items and 6 identical items in the mse indicator, with only 3 items worse than PatchTST. In the mae indicator, UnetTSF has 10 leading items, 1 identical item, and 4 items worse than PatchTST

\textbf{Multivariate Time-series Forecasting}:Table \ref{tab:multivarite} summarize the multivarite evaluation results of all the methods on eight datasets. We cite the baseline results from $ (Zeng et al., 2022) $.Overall, the UnetTSF model performed very well in testing. Among the 32 tests on 8 datasets, 10 achieved the best results in terms of mse and 15 achieved the best results in terms of mae.Compared to DLinear, the average decrease in mse was 10.1\% and the decrease in mae was 9.1\%. On the most periodic dataset weather, we achieved all optimal results, with UnetTSF and patchTST showing a decrease of 1.0\% in mse and 0.4\% in mae, and a decrease of 12.6\% and 15.1\% in mae compared to DLinear. On the traffic and electricity datasets, UnetTSF showed a decrease of about 4\% in mse compared to DLiner. On the ILI dataset, UnetTSF achieved the best results in predicting a length of 60. Compared with PatchTST, mse decreased by 7.4\% and mae decreased by 6.4\%. Compared with DLinear, mse decreased by 66.6\% and mae decreased by 46.7\%. Among the 16 tests on the ETT dataset, UnetTSF achieved 11 optima in mae and mse achieved 4 optima,Especially on the test item with a predicted length of 192 on ETTh1, UnetTSF showed a 3.4\% decrease in mse and a 3.5\% decrease in mae compared to PatchTST.

\textbf{model efficiency}:Table \ref{tab:parameter} The parameter quantity and computational complexity are one of the important evaluation indicators of the model.We used a fixed length $L=336$ review window, a predicted future length $T=96$, and a $batch=32$ configuration on the ETTh2 dataset to evaluate the parameter count, computational complexity, and GPU memory usage of all model inference stages.In Tab \ref{tab:parameter} we compare the average practical efficiencies with 5 runs.
\begin{itemize}
	\item In terms of computational complexity, DLiner and UnetTSF have significant advantages. PatchTST uses patch technology to significantly reduce computational complexity, but it is still 12.1 times that of UnetTSF. UnetTSF and DLiner both use fully connected and pooling layers, while UnetTSF uses a small pooling kernel to reduce computational complexity by 6.6\%.
	\item In terms of model parameter quantity and inference memory usage, UnetTSF has the same magnitude as DLline and PatchTST. Combining the results of univariate and multivariate time series prediction, UnetTSF is more suitable for use in limited resource scenarios than DLline and PatchTST.
\end{itemize}
\section{Conclusion}
This paper proposes a long time series prediction model UnetTSF with linear complexity. UnetTSF has two innovative points: the first proposes an FPN description structure for time series data, which becomes the third choice for raw data description and binary decomposition methods for time series data. The second is to introduce the Unet network structure into the field of time series prediction, and design and implement an Unet network suitable for time series prediction. The experimental results show that UnetTSF has superior performance compared to DLiner and PatchTST, and UnetTSF with linear complexity is more suitable for practical production and life.

\end{document}